\documentclass{article}
\usepackage{ijcai22}

\usepackage{times}

\usepackage{soul}
\usepackage{url}
\usepackage[hidelinks]{hyperref}
\usepackage[utf8]{inputenc}
\usepackage[small]{caption}
\usepackage{graphicx}
\usepackage{amsmath}
\usepackage{booktabs}
\usepackage[T1]{fontenc}
\usepackage{tikz}
\usetikzlibrary{automata, positioning, arrows.meta}

\usepackage{amsthm}
\usepackage{amsfonts}
\usepackage{caption,subcaption}

\usepackage[sorting=none]{biblatex}
\theoremstyle{plain}
\newtheorem{theorem}{Theorem}

\theoremstyle{definition}
\newtheorem{definition}[theorem]{Definition}

\usetikzlibrary{calc}

\title{Equivariant neural networks for recovery of Hadamard matrices}

 \author{Augusto Peres$^1$\and
   Eduardo Dias$^1$\and
   Luís Sarmento$^1$\And
   Hugo Penedones$^1$\\
   \affiliations
   $^1$Inductiva Research Labs\\
   \emails
   \{augusto.peres, eduardo.dias, sarmento, hpenedones\}@inductiva.ai
 }

 \date{
 	Inductiva Research Labs \\ \texttt{\{augusto.peres, eduardo.dias,
     sarmento, hpenedones\}@inductiva.ai}
}

\begin{document}

\maketitle
	
\begin{abstract}
  We propose a message passing neural network architecture designed to be
  equivariant to column and row permutations of a matrix. We illustrate its
  advantages over traditional architectures like multi-layer perceptrons (MLPs),
   convolutional neural networks (CNNs) and even Transformers, on the
  combinatorial optimization task of recovering a set of deleted entries of a
  Hadamard matrix. We argue that this is a powerful application of the
  principles of Geometric Deep Learning to fundamental mathematics, and a
  potential stepping stone toward more insights on the Hadamard conjecture using
  Machine Learning techniques.
\end{abstract}


\section{Introduction}

Hadamard matrices are matrices whose entries are either $-1$ or $+1$ and their
rows are mutually orthogonal. A necessary condition for a $\{-1, 1\}$-matrix to
be Hadamard is being of order either $n = 2$ or $n = 4k$ for $k \in \mathbb{N}$.
There are known examples of Hadamard matrices for many orders $4k$ (see for
example~\cite{txtHad}), as well as for all orders where $n = 2^k$, but finding
or generating an example of a Hadamard matrix for every $n = 4k$ remains an open
challenge. In fact, proving that there are examples of Hadamard matrices for
every $n = 4k$ (even if we cannot generate them) is a long-standing conjecture
in Mathematics named \textit{Hadamard’s conjecture}. At the time of writing the
smallest integers for which no Hadamard matrix is yet known are $n = 668, 716
\textup{ and } 892$.

Interestingly, challenges related to the Hadamard conjecture come at even
smaller orders. For example, it is also conjectured that \textit{symmetric}
Hadamard matrices should exist for every order $4k$. However, not a single
symmetric Hadamard matrix of order 236 is known, and it was only very recently
that symmetric Hadamard matrices of sizes as small as 188, 172 or even 116 were
finally found (see~\cite{symmetric47_73_113} and~\cite{OliviaDiMatteo2015}).

Since any operation involving permutations or negations of rows or columns of an
Hadamard matrix do not affect the orthogonality of rows / columns, finding an
example of a Hadamard matrix implies finding an entire class of
\textit{equivalent matrices}. Two Hadamard matrices $H$ and $H'$ are equivalent
if one can be obtained from the other by row/column permutations and row/column
negations. The set of all matrices that are equivalent to $H$ is called the
\textit{equivalence class} of $H$, and any matrix in a given equivalence class
is referred to as a representative of that class. Finding a representative of
each equivalence class for all $4k$ orders is still a harder problem, and it is
still open even for relatively small orders. Equivalence classes of Hadamard
matrices have only been fully enumerated up to size 32, of which there exist
13710027, see~\cite{equivalenceClasses}.

The search for Hadamard matrices has mostly relied on human intuition and
expertise. Traditional methods, such as Paley’s
construction~\cite{paleyConstruction}, Sylvester’s
construction~\cite{sylvester67}, Williamson’s method~\cite{williamsons} and
several others~\cite{browne2021survey} rely on a constructive approach.
Typically, these consist in assuming that a Hadamard matrix can be constructed
from a number of smaller block matrices using a simple concatenation procedure.
The challenge then becomes that of finding such smaller blocks, and the hope is
that by imposing a number of restrictions over the structure of such blocks,
that search becomes easier. However, such methods are not general, and are only
applicable to a subset of all possible orders. For example, Sylvester’s
construction can only be applied to $n = 2^k$.

One way to avoid hand-crafted constructive methods in the search for Hadamard
matrices is to formulate the problem as a combinatorial optimization instance.
Then, using the vast arsenal of computational tools made available in the last
decade such as constraint satisfaction programming (CSP), answer set
programming, genetic algorithms, and a number of search algorithms
(\textit{e.g.} hill climbing) search for solutions. However, due to the enormous
size of the solution space, classical methods have not yet been able to find
solutions to small instances. For example,
in~\cite{simmulatedSpinnVectors,e20020141} the authors found Hadamard matrices
using variations of simulated annealing but where only able to scale up to
matrices of order 20. Our own experimentation with CSP systems has only allowed
us to replicate known results up to size 28. It seems that progress in this
front requires devising heuristics capable of dramatically restricting the
search space, but that is in itself a huge mathematical challenge that requires
deep expertise.

Recently, there have been efforts in applying Machine Learning (ML) methods to
tackle combinatorial optimization problems that classical algorithms struggle
with. As stated in~\cite{bengio2020machine}, Deep Learning (DL) could be used in
two ways. Firstly, we can replace computationally expensive heuristics by their
approximations. Such is the case in~\cite{agostinelli2021a} where neural
networks are used to replace computationally expensive heuristics in the
$A^{\star}$ algorithm. Alternatively, when no heuristics have yet been
developed, we can use DL to explore the space of possible decisions gathering
experience until a proper policy regarding which feasible solutions to explore
has been achieved. This approach has been successful in~\cite{NIPS2017_d9896106}
where graph neural networks are trained to solve combinatorial optimization
problems over graphs and in~\cite{kurin2020qlearning}, where neural networks
(NN) are used to make better variable assignments in boolean satisfiability
(SAT) instances.


Our work follows this recent trend of approaching combinatorial problems using
DL methods. However, at the time of writing, there has been little work on DL
related to any type of matrix operations and information regarding which
architectures and techniques work well is scarce.


Therefore, as a first attempt, we opt for a more modest, approach. Specifically,
we focus on the \textit{matrix completion task}. Here a random number of the
$\{-1, 1\}$ entries of Hadamard matrices $H$ is set to zero. Then, we want to
train NNs to reconstruct the matrices by predicting the original value of each
zeroed-out entry.




Previous work on reconstruction without using DL has been established
in~\cite{KLINE201933}, where the author develops an algorithm to recover
uniformly-distributed zeroed-out entries by using simple matrix operations. The
author then integrates this recovery algorithm in an optimization procedure to
generate Hadamard matrices.

In this paper, we investigate whether or not the results of the algorithm
designed in~\cite{KLINE201933} can be reproduced by simply letting neural
networks (NN) see examples of matrices with zeroed-out entries and their
respective reconstructions. However, this tasks is hard for modern NN since they
do not have any inductive bias on how row and columns permutations to the input
should affect the output.

Clearly, if we move a given row that requires completion to another place in the
matrix those values will remain unchanged, only their position was affected.
This is addressed in another key contribution of our work, where a model with
this inductive bias is constructed. Lastly we will explore some properties of
the trained models. More specifically how they can be used to slow the
performance degradation of the completion algorithms that happens naturally as
more and more entries are zeroed-out.

\section{Related work}

Hadamard matrices have received a lot of attention since the conjecture was
first put forward with early works focused on constructive
methods~\cite{paleyConstruction,williamsons,sylvester67} for the generation of
matrices. More recent approaches have resourced to computational methods for
finding them. For instance, in~\cite{doi:10.1080/09720529.2005.10698034} a
genetic algorithm, using algebraic concepts for the backbone of the fitness
function, yielded matrices of orders 52, 56, 60, 64.
In~\cite{e20020141,simmulatedSpinnVectors} the authors use simulated annealing
to find Hadamard matrices but unfortunately can only scale up to size 20.
Reconstruction of Hadamard matrices was recently addressed
in~\cite{Goldberger2021} where, instead of recovering randomly placed zeroed-out
entries, the authors propose an algorithm to create Hadamard matrices starting
with just the first $n/2$ rows.

As previously mentioned, in~\cite{KLINE201933} a method capable of recovering
$\mathcal{O}(n^2/\log(n))$ entries is presented. It completes a matrix $X$ with
zeroed out entries using the expression $X - \textup{sign} (t_s(X^t - nX^{-1}))$
where $t_s$ sets all entries smaller than $s$ to zero. This method is then used
to yield Hadamard matrices of several sizes and one symmetric Hadamard matrix of
size $116$ which had only recently been discovered.

Here we give the first steps towards improving their bound. We stress that the
reconstruction task is hard for NNs due to the large state space and
how row/column permutations/negations affect the input.

As previously mentioned, in the reconstruction task, if a permutations is
applied to the input then that same permutation should be applied to the target
predictions. Constructing NNs that have the geometries that act on
the inputs (such as graphs, sets, grids, geodesics and meshes) imprinted on
their structure is the target study of what is known as \textit{Geometric Deep
  Learning} (GDL)~\cite{bronstein2021geometric}.

Message passing neural networks, which are specifically designed to work with
graph-like structures as inputs, have recently surpassed state-of-the-art
results in molecule prediction benchmarks~\cite{gilmer2017neural}.
In~\cite{barrett2020exploratory}, message passing neural networks are used to
parameterize node selection policies learned through reinforcement learning to
solve the Max-Cut problem. The learned policies surpassed previous methods, such
as the one presented in~\cite{NIPS2017_d9896106}, which did not use message
passing neural networks. In~\cite{NIPS2017_f22e4747} models were designed to
approximate functions defined over sets that are invariant to permutations and
achieve state-of-the-art performance in digit summation and set expansion.


There is also a growing body of literature focusing on integrating the existing
examples of GDL, such as the ones referenced above, into a
general framework from which they are simpler particular instances. Such is the
case of~\cite{bronstein2021geometric}, where they present a blueprint for
building equivariant models under the actions of any symmetry group acting on
the inputs of the NN. They also collect examples from previous works
and show how they fit in their blueprint. In~\cite{ravanbakhsh2017equivariance}
the authors show how equivariant NNs can be built through parameter
sharing and they conclude that a layer is equivariant to the actions of some
group if and only if that group is the symmetry group of the bipartite graph
representing the layer.

In~\cite{kondor2018generalization} the authors prove that convolutional
structure is both necessary and sufficient for a neural network to be
equivariant to the actions of some compact group. This result is then used
in~\cite{thiede2020general} to create two distinct layers equivariant to
permutations. The first receives arrays as input and is equivariant to
permutations of the elements. The second receives matrices as input and is
equivariant to permutations acting on rows and columns. This differs from our
model because, while in ours, different permutations can be applied to the rows
and columns, in theirs the same permutation must be applied to the rows and
columns.

Despite only recently being formalized in scientific
literature~\cite{bronstein2021geometric}, the concepts of geometric deep
learning, have been used for quite some time. For example, convolutional neural
networks (CNNs), which capture translation equivariance, have been used since
the 80's~\cite{10.1162/neco.1989.1.4.541} in computer vision tasks with a much
greater deal of success than multi-layer perceptrons which are not endowed with
such inductive bias. Recently, CNNs have been generalized to arbitrary symmetry
groups such as rotations and reflections~\cite{NIPS2014_f9be311e,
  cohen2016group}. Both these methods achieved state of the art performance on
the rotated MNIST.\@

\section{Implementing equivariance}

In the completion task, \textit{i.e}, predicting the values of zeroed-out matrix
entries, if we apply to distinct permutations to the rows and columns of $M$,
then the same permutations should be applied to the target predictions for the
missing values. This is known as equivariance.

More formally, this is known as equivariance to the actions of the group
$S_n^2$. Formal definition of \textit{equivariance}, \textit{invariance},
$S_n^2$ and how actions of $S_n^2$ affect squared matrices as well as a brief
introduction to groups is given in Appendix~\ref{sec:grou-appendix}. In this
section we are going to build two equivariant layers to the actions of $S_n^2$
($S_n^2$-equivariant).

A simple way to construct an $S_n^2$-equivariant layer is to apply the same
function $\psi$ entry-wise to matrix $M$. More formally, if we denote $M$ as the
input to this layer and $M'$ as its output, then the entry $(i, j)$ of $M'$ will
be given by
\begin{align}
  \label{eq:final_layer_expression}
  M'_{(i, j)} = \psi(M_{(i, j)})
\end{align}


Equivariance is maintained even if we consider each entry in $M$ to be an array
of features of dimension $k$ instead of real numbers, when this happens we say
that $M \in \mathbb{R}^{n^2k}$. Of course in that case $\psi$ would have to map
$\mathbb{R}^k$ into $\mathbb{R}^{k'}$.

This layer, despite being equivariant, is not very expressive since $M'_{(i,
  j)}$ will have only used local information, \textit{i.e}, information from
$M_{(i, j)}$. Success in the reconstruction tasks requires gathering global
information, \textit{i.e}, information about the whole matrix.

To achieve this while maintaining equivariance we use a mechanism of message
passing similar to the one used on graph nets~\cite{gilmer2017neural}. Our
layer, in order to compute $M'_{(i, j)}$, will use the information present on
$M_{(i, j)}$, its row and its column.



We first describe how we use the information present in the $i$-th row for the
computation of $M'_{(i, j)}$. A key observation is that this computation must be
invariant to column/row permutation. We do this by pairing $M_{(i, j)}$ with
every other element in its row and then feed all pairs to a function $\psi_r$.
The $n-1$ results are aggregated using a permutation invariant operation. We
denote this as $\Psi_{\textup{r}}(i, j)$, given by:
\begin{align}
  \label{eq:row-information-extraction}
  \Psi_{\textup{r}}(i, j) = \bigoplus_{w \in \{1, \dots, n\} \setminus \{j\}} \psi_{\textup{r}}(M_{(i, j)} , M_{(i, w)})
\end{align}
where $\bigoplus$ is some permutation invariant aggregator such as $\sum$ or
$\prod$.


Using the information present in the $j$-th column for the computation of
$M'_{(i, j)}$ follows the same procedure:
\begin{align}
  \label{eq:col-information-extraction}
  \Psi_{\textup{c}}(i, j) = \bigoplus_{w \in \{1, \dots, n\} \setminus \{i\}} \psi_{\textup{c}}(M_{(i, j)} , M_{(w, j)})
\end{align}

To complete our \textit{equivariant message passing layer} (EMP) we simply
aggregate $\Psi_{\textup{r}}$ and $\Psi_{\textup{c}}$ using any invariant
operator $\oplus$. More formally, let $M'$ denote the output of the layer then
$M'_{(i, j)}$ is given by the equation:
\begin{align}
  \label{eq:layer-formula}
  M'_{(i, j)} = \Psi_{\textup{r}}(i, j) \oplus \Psi_{\textup{c}}(i, j)
\end{align}

In order to make accurate predictions in the reconstruction task, we need to
gather information about the whole matrix, yet here we are only using, for each
element of the matrix, elements in its row and column. However notice that if we
compose two EMPs $l_1$ and $l_2$ then each entry of $l_2(l_1(M))$ will have
gathered information about every other entry in the matrix.

In fact this is the minimum~\textit{degree of connectivity} that is required to
propagate information about the whole matrix equivariantly. If we were to use
only information in rows then the elements in the matrix would be blind to any
element not on its row regardless of the number of layers stacked together. If
we were to use only part of the columns/rows then the layer would not be
equivariant. If we were to simply combine $M_{(i, j)}$ with all other elements
in the matrix and then feed them to $\psi$ the layer would be equivariant, it
would gather information about the whole matrix, but it would be too
computationally expensive.

\begin{figure}[h]
  \centering
  
  \begin{tikzpicture}[
    >=Stealth,
    shorten >=1pt,
    auto,
    node distance=1 cm,
    scale = 1,
    transform shape,
    state/.style={rectangle,inner sep=2pt}]

    \node[state] (q2) [] {$\begin{bmatrix}
        (a) & (b)\\
        (c) & (d)
      \end{bmatrix}$};

    \node[state] (q3) [below=of q2] {$\begin{bmatrix}
        \psi_{\textup{r}}(a, b) + \psi_{\textup{c}}(a, c) & \psi_{\textup{r}}(b, a) + \psi_{\textup{c}}(b, d)\\
        \psi_{\textup{r}}(c, d) + \psi_{\textup{c}}(c, a) & \psi_{\textup{r}}(d, c) + \psi_{\textup{c}}(d, b)
      \end{bmatrix}$};

    \node[state] (q4) [below= of q3] {$\vdots$};

    \node[state] (q5) [below= of q4] {$\begin{bmatrix}
        \mathbf{x}_1 \in \mathbb{R}^k & \mathbf{x}_2 \in \mathbb{R}^k\\
        \mathbf{x}_3 \in \mathbb{R}^k & \mathbf{x}_4 \in \mathbb{R}^k
      \end{bmatrix}$};

    \node[state] (q6) [below=of q5] {$\begin{bmatrix}
        x_1 \in [-1, 1] & x_2 \in [-1, 1]\\
        x_3 \in [-1, 1] & x_4 \in [-1, 1]
      \end{bmatrix}$};

    \path[->]
    (q2) edge [left] node {EMP} (q3)
    (q3) edge [left] node {EMP} (q4)
    (q4) edge [left] node {EMP} (q5)
    (q5) edge [left] node {Entry wise classifier} (q6);
  \end{tikzpicture}
  \caption{Scheme of an equivariant model achieved by composition of several
    equivariant layers.}
  \label{fig:model-usage}
\end{figure}
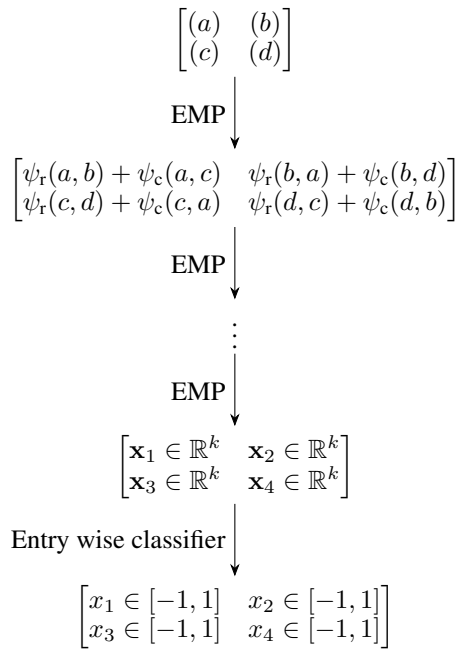

To create an equivariant model (EMPM) we simply stack several EMPs together and
then add a final layer following equation~\ref{eq:final_layer_expression} as
it is illustrated in Figure~\ref{fig:model-usage}.


\section{Data generation and training}

Our goal in the next sections is to show that DL methods are able to learn
Hadamard matrices reconstructions and that, by using GDL
techniques, we can effectively deal with the large state space arising from row
and column permutations.

This is done over two different experiments. Firstly we train and evaluate the
performance of our models inside a single equivalence class. Secondly we train
and evaluate our models in different classes.




For our experiments we want data sets consisting on several examples of Hadamard
matrices with zeroed-out entries and their respective reconstruction. That is,
our data set will contain tuples of the form:
\begin{align}
  \left(\begin{bmatrix}
      1 & 1 & 1 & 0 \\
      1 & -1 & 1 & -1 \\
      1 & 1 & -1 & -1 \\
      1 & 0 & -1 & 1 \\
    \end{bmatrix} , \begin{bmatrix}
      0 & 0 & 0 & 1 \\
      0 & 0 & 0 & 0 \\
      0 & 0 & 0 & 0 \\
      0 & -1 & 0 & 0 \\
    \end{bmatrix}\right)
\end{align}

To create the data sets we first gathered Hadamard matrices from~\cite{txtHad},
where examples of matrices for several different sizes can be found, and one
representative of every equivalence class up to size 28 is available. This is
actually the only size for which we test generalization across classes. Ideally
we would have liked to test generalization to unseen classes for $n=32$, the
last integer for which all classes are known, but there were not enough
representatives of this order in~\cite{txtHad} for meaningful experimentation.

For sizes 8, 12, 16, 20 and 24 there exist only 1, 1, 5, 3 and 60 different
equivalent classes, respectively, so we decided to skip those orders as well.

When training models to generalize to unseen classes, we split the
representatives of equivalent classes into two disjoint sets. One of those sets
will be used to generate training examples and the other will be used to
generate the validation examples. For all other experiments we use the same
representative to generate the training and validation data. Each element in the
data set is generated as follows:

\begin{itemize}
\item From the available training/validation representatives we randomly select
  one matrix $H_n$.
\item If the model we are training is not $S_n^2$-equivariant we introduce a
  random number of row/column permutations in $H_n$. This is simply a data
  augmentation strategy.
\item Next we negate random rows/columns. This data augmentation is needed
  because the models do not exploit any geometric properties related to
  negations.
\item Next we randomly choose entries in $H_n$ to be set to zero. The values of
  those entries will be stored in the target matrix in their respective
  positions.
\end{itemize}

\section{Results}

\subsection{EMPM vs other baselines}

We start this section by showing that the EMPM presented in this paper has a
better performance on the reconstruction task than the other models tested. This
comparison is necessary since our goal is to actually use the models to
replicate the reconstruction used in~\cite{KLINE201933}, henceforth referred to
as Kline's method. Therefore, before moving forward, we should know which models
are better suited for said task.

The way we use our models to reconstruct the matrices is straightforward. We
simply give them Hadamard matrices with zeroed out entries and allow them to
make a prediction. Each entry is then completed with the sign of its
correspondent prediction. More formally, let $H$ and $m$ the the matrix with
zeroed out entries and $m$ some model, then $H$ is decoded into $H +
\textup{sign}(m(H))$. We call this~\textit{one-shot matrix reconstruction}.

To compare the EMPM to other models we also trained MLPs, CNNs and a
transformer-like architecture. From these three baseline models the CNN yielded
the best results therefore we leave the details regarding the other
architectures for the appendix.

Table~\ref{tab:comparisson} shows the performance of the EMPM when compared to a
CNN in the one-shot reconstruction task. As we can see the EMPM outperforms the
CNN for any given number of zeroed-out entries. Moreover, unlike the CNN, the
EMPM shows no degradation when we move from $n=8$ to $n=12$.

Strangely, several times the performance of the EMPM for $n=12$ was better than
for $n=8$. We think that this is easily explained given the stochastic nature of
the reconstruction.

To mitigate the degradation of the CNN at $n=12$ we increased the size of the
model but no better results where obtained. This contrasts with the EMPM where
the exact same network structure was used from $n=8$ to $n=32$ without much
degradation to its performance. More details on both architectures in the
appendix.

\begin{table*}[h]
  \centering
  \begin{tabular}{|l|l|l|l|l|}
    \hline
    Zeroed-out entries & EMPM ($n=8$) & CNN ($n=8$) & EMPM ($n=12$) & CNN ($n=12$) \\ \hline
    1                  & 1            & 0.883        & 1.0           & 0.743         \\ \hline
    2                  & 1.0          & 0.791        & 0.999         & 0.574         \\ \hline
    3                  & 0.998        & 0.65         & 0.999         & 0.405         \\ \hline
    4                  & 0.993        & 0.54         & 0.98          & 0.3           \\ \hline
    5                  & 0.964        & 0.423        & 0.961         & 0.199         \\ \hline
    6                  & 0.872        & 0.316        & 0.918         & 0.143         \\ \hline
    7                  & 0.776        & 0.241        & 0.843         & 0.103         \\ \hline
    8                  & 0.614        & 0.16         & 0.765         & 0.079         \\ \hline
  \end{tabular}
  \caption{Percentage of successful one-shot matrix reconstructions as a
    function of the zeroed out entries. The first column corresponds to the
    number of zeroed out entries. The reaming columns correspond to the success
    rates of the models on the one-shot reconstruction task. For brevity we
    decided to show only the number of zeroed out entries until $8$.}
  \label{tab:comparisson}
\end{table*}

\subsection{EMPM vs Kline}

In this section we compare the EMPM reconstruction capabilities with Kline's
method.

\begin{figure*}[htb]
  \centering
\begin{subfigure}{0.33\textwidth}
    \includegraphics[width=\linewidth]{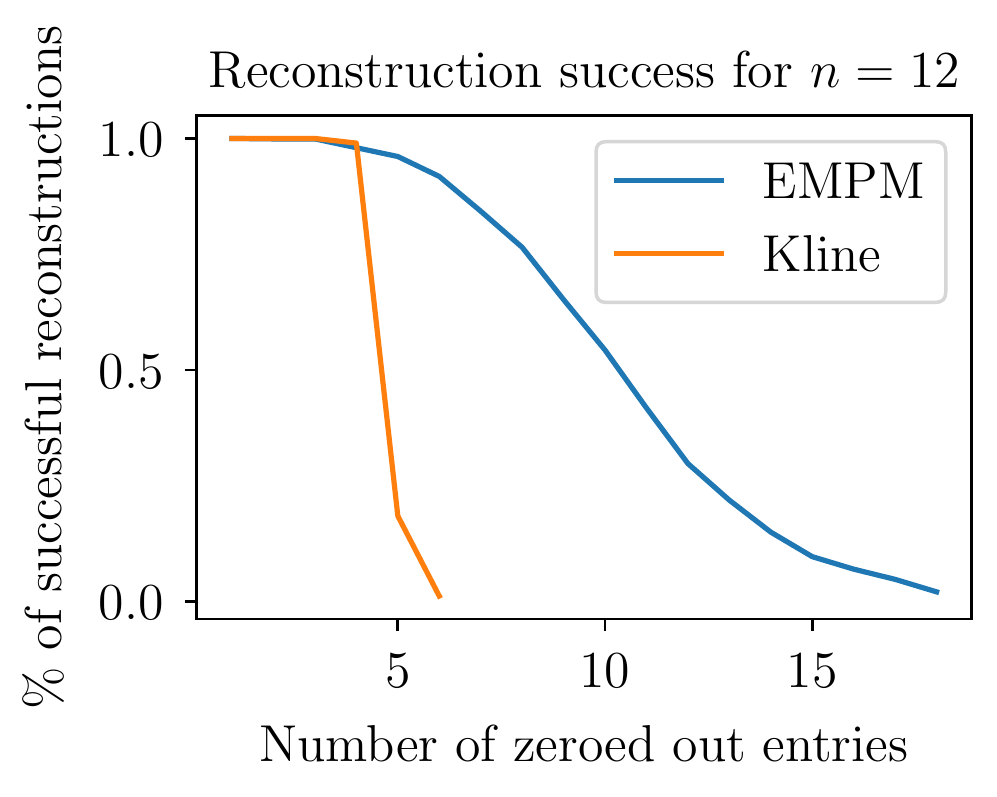}
    \caption{$n=12$}\label{fig:kline-comparisson-size-12}
\end{subfigure}\hfil
\begin{subfigure}{0.33\textwidth}
  \includegraphics[width=\linewidth]{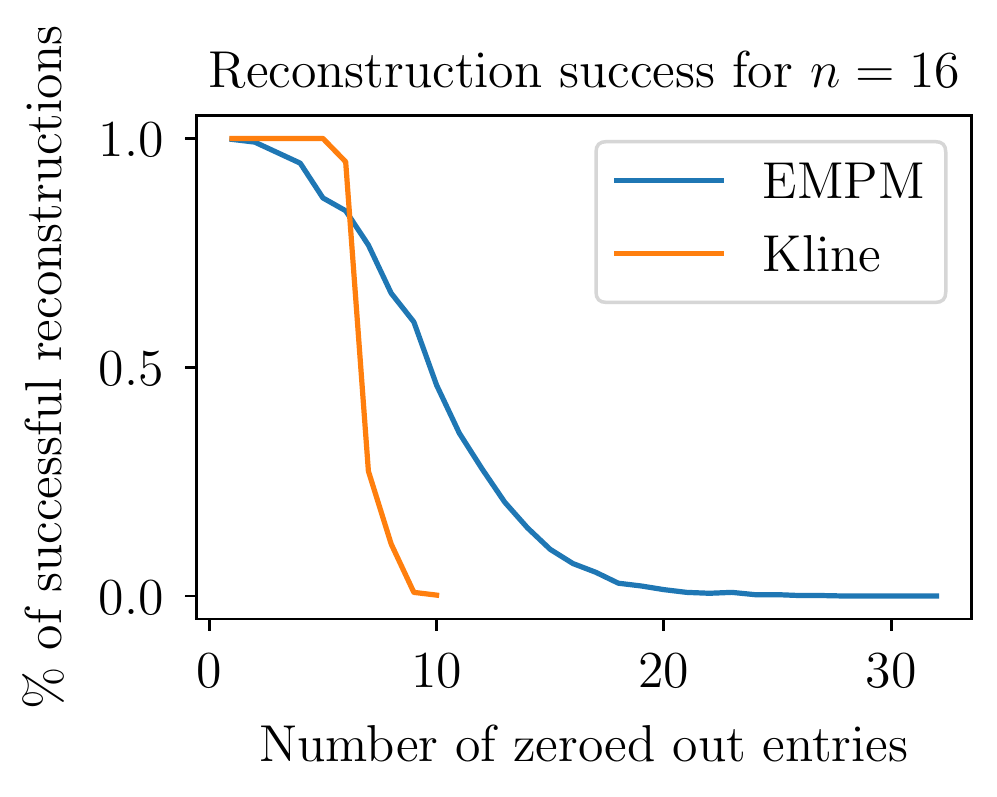}
  \caption{$n=16$}\label{fig:kline-comparisson-size-16}
\end{subfigure}\hfil
\begin{subfigure}{0.33\textwidth}
  \includegraphics[width=\linewidth]{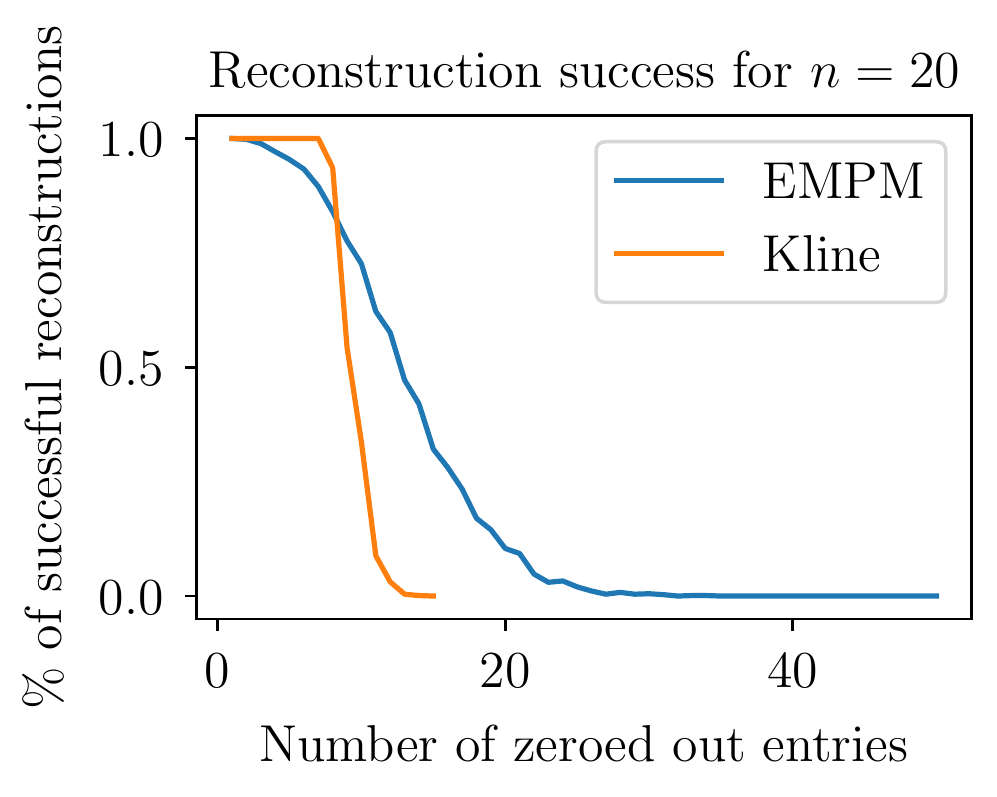}
  \caption{$n=20$}\label{fig:kline-comparisson-size-20}
\end{subfigure}

\medskip
\begin{subfigure}{0.33\textwidth}
  \includegraphics[width=\linewidth]{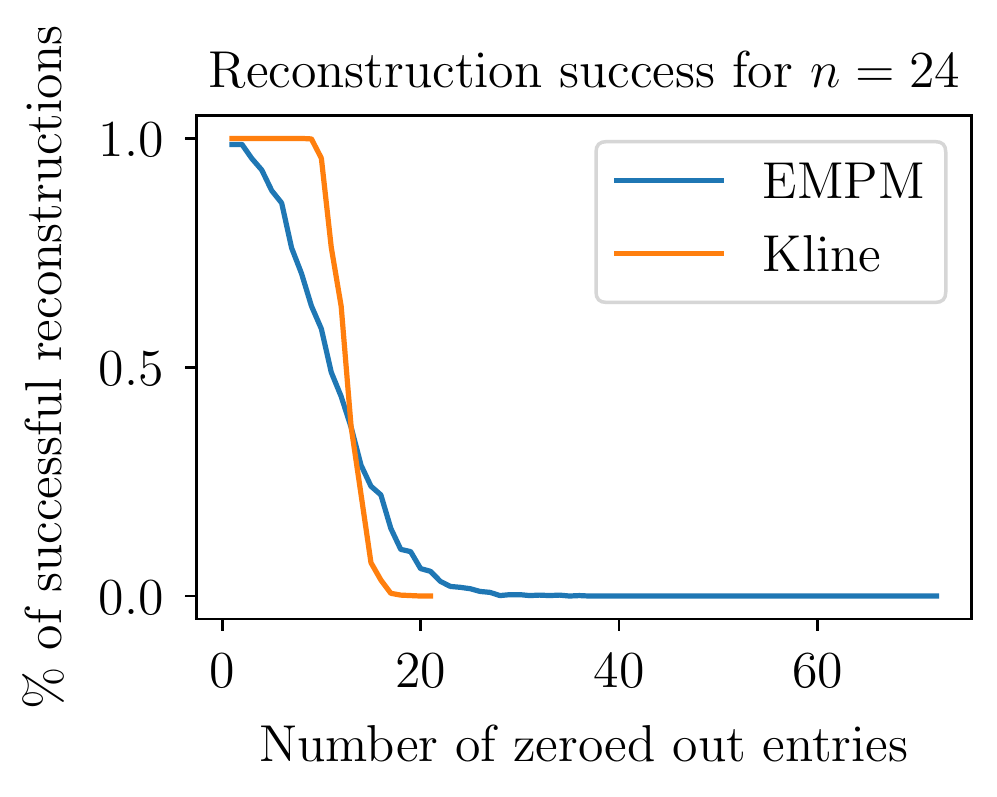}
  \caption{$n=24$}\label{fig:kline-comparisson-size-24}
\end{subfigure}\hfil
\begin{subfigure}{0.33\textwidth}
  \includegraphics[width=\linewidth]{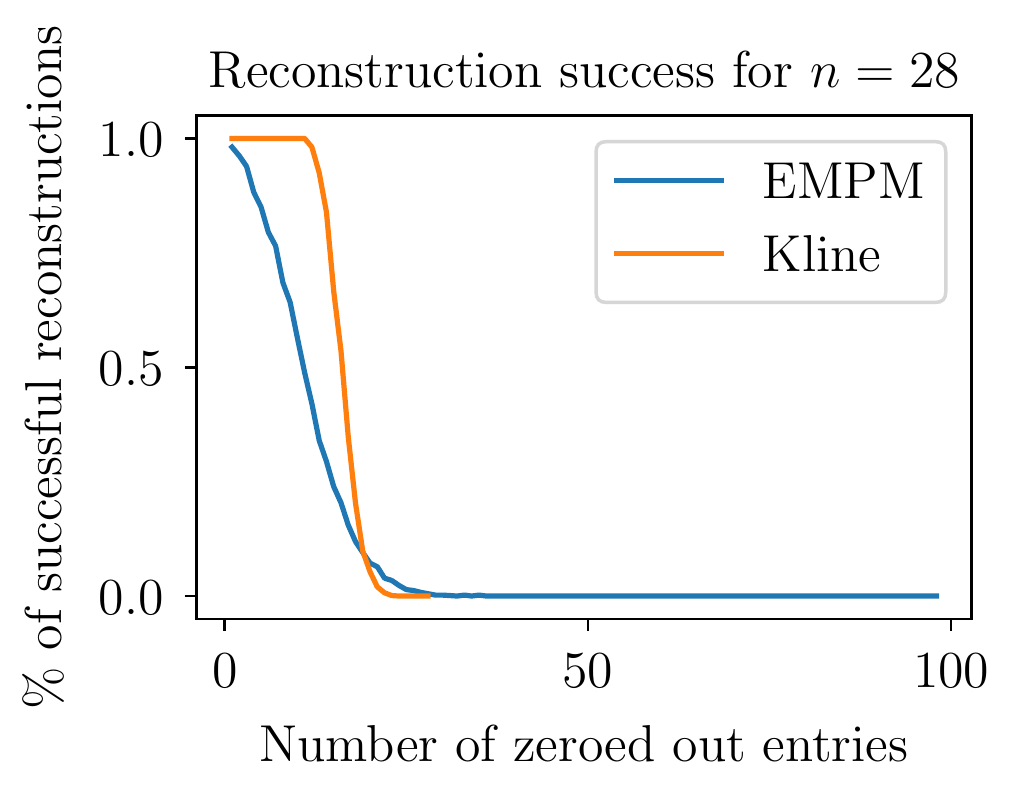}
  \caption{$n=28$}\label{fig:kline-comparisson-size-28}
\end{subfigure}\hfil
\begin{subfigure}{0.33\textwidth}
  \includegraphics[width=\linewidth]{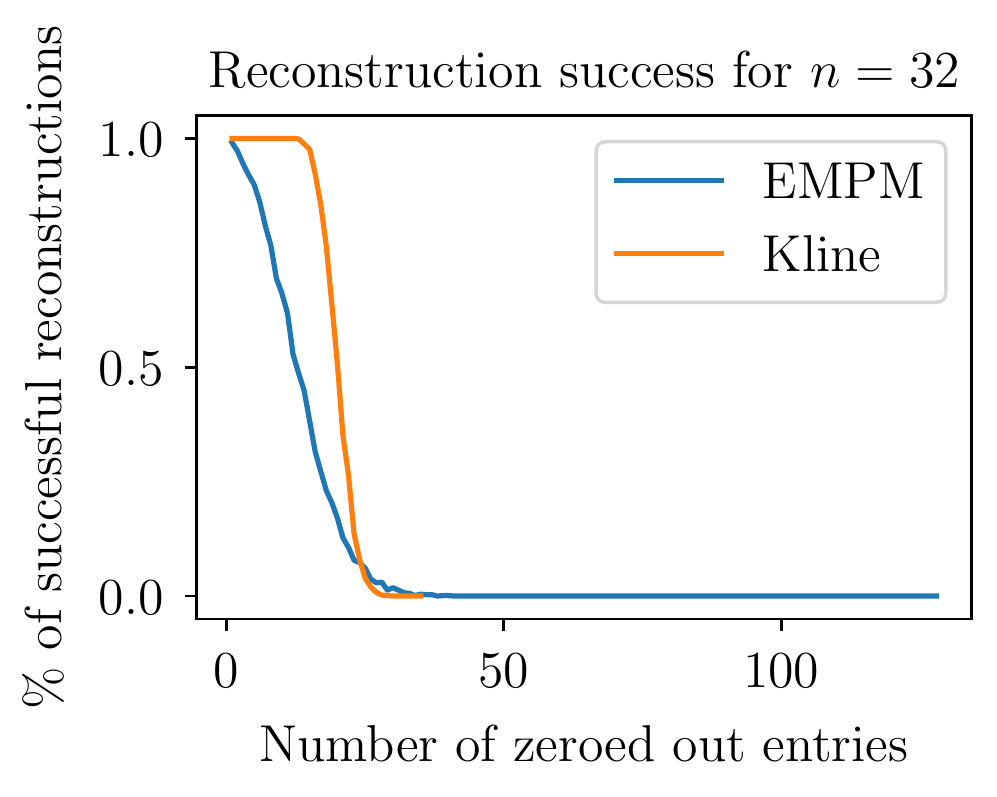}
  \caption{$n=32$}\label{fig:kline-comparisson-size-32}
\end{subfigure}
\caption{Percentage of successful reconstructions obtained by both the EMPM
  (blue) and Kline's method (yellow). The $x$ axis shows the number of zeroed
  out entries. The $y$ axis shows the percentage of successful matrix
  reconstructions.}
\label{fig:kline-comparisson-all}
\end{figure*}

Figure~\ref{fig:kline-comparisson-all} shows the percentage of successful
reconstructions for both Kline's method and the EMPM as a function of the number
of zeroed out entries. We can see that, until $n=20$, our method outperforms
Kline's. However for $n \geq 24$ the EMPM is outperformed.

Nevertheless we can see that both methods enter the region where virtually no
matrix is reconstructed at about the same number of zeroed-out entries. And,
still inside the domain where both methods are able to recover matrices, the
EMPM has a good performance when compared to what is currently, to the extent of
our knowledge, the best available method for Hadamard matrix reconstruction in
scientific literature~\cite{KLINE201933}.

At the moment we believe that the EMPM is outperformed mainly due to the
stochastic nature of the decoding, \textit{i.e}, since our model is predicting
the values of missing entries it is expected that, with some probability, it
will eventually miss. Of course the more entries the model has to predict, the
greater the chance for error and the less the chance for success. Klines's does
not suffer from this as no probabilistic predictions are required for the
reconstruction.

\subsection{Generalization to other equivalence classes}

The results shown so far have all been obtained by training and evaluating the
model over the same equivalence class. However, if we ever hope to use machine
learning to find Hadamard matrices, the ability to generalize to unknown
equivalence classes is a must.

\begin{figure}[h]
  \centering
  \includegraphics[width=0.40\textwidth]{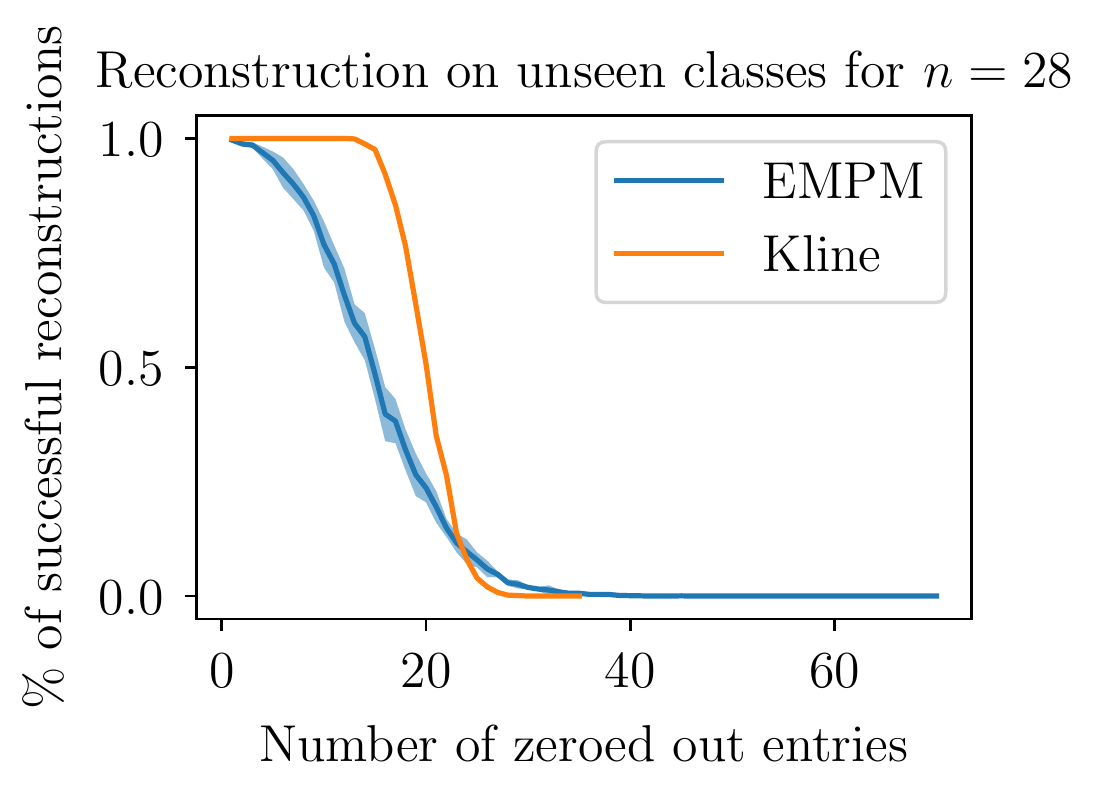}
  \caption{Percentage of reconstruction success for a model trained on 24 of the
    487 equivalence classes for $n=28$ and evaluated on the remaining ones. The
    filled region corresponds to the 95\% confidence interval of the
    mean.}\label{fig:size_28_train_24}
\end{figure}

Figure~\ref{fig:size_28_train_24} compares the reconstruction capabilities of
the EMPM with Kline's method. This time, the model was trained on 24,
corresponding to 5 percent, of the 487 existing equivalence classes for $n=28$.
We performed five different experiments where the matrices used for training
were chosen randomly at the start of each experiment. The averages of both
metrics across all experiments were taken to draw the darker lines in the plot.
The filled region corresponds to the 95\% confidence interval of the mean,
\textit{i.e}, the region $\overline{x} \pm 1.96 \sigma/\sqrt{5}$, where $\sigma$
is the standard deviation and $\overline{x}$ is the mean of the percentage of
successes for each number of zeroed-out entries.

As we can see by comparing Figures~\ref{fig:kline-comparisson-size-28}
and~\ref{fig:size_28_train_24} our model does not lose reconstruction
capabilities when evaluated on unseen equivalence classes. Moreover we can also
conclude that we require very few representatives of distinct equivalence
classes to achieve generalization to the remaining classes.

\subsection{Extending the performance of the EMPM}

When evaluating our models we created a softer metric called \textit{highest
  confidence prediction} (HCP). Here we check if the sign of the entry in the
model's prediction with the largest absolute value corresponds to the sign of
the ground truth.

\begin{figure}[h]
  \centering
  \includegraphics[width=0.40\textwidth]{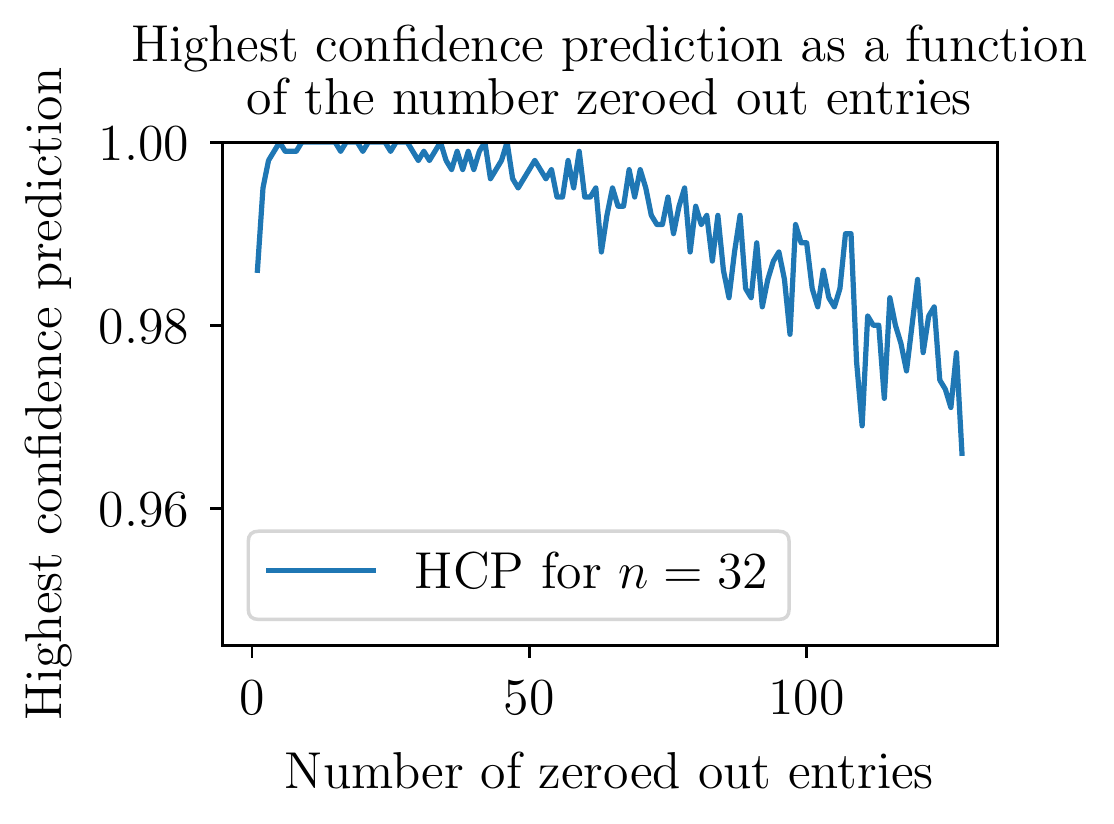}
  \caption{Highest confidence prediction as a function of the number of
    zeroed-out entries.}
  \label{fig:highest-confidence-prediction}
\end{figure}

When we evaluated this metric as a function of the number of zeroed-out entries
we observed that it suffered from very little degradation in performance as the
number of zeroed out entries increased. This is illustrated
in~\ref{fig:highest-confidence-prediction} where the metric is computed for
$n=32$ with the number of zeroed-out entries ranging from $1$ to $128$.

\begin{figure}[h]
  \centering
  \includegraphics[width=0.40\textwidth]{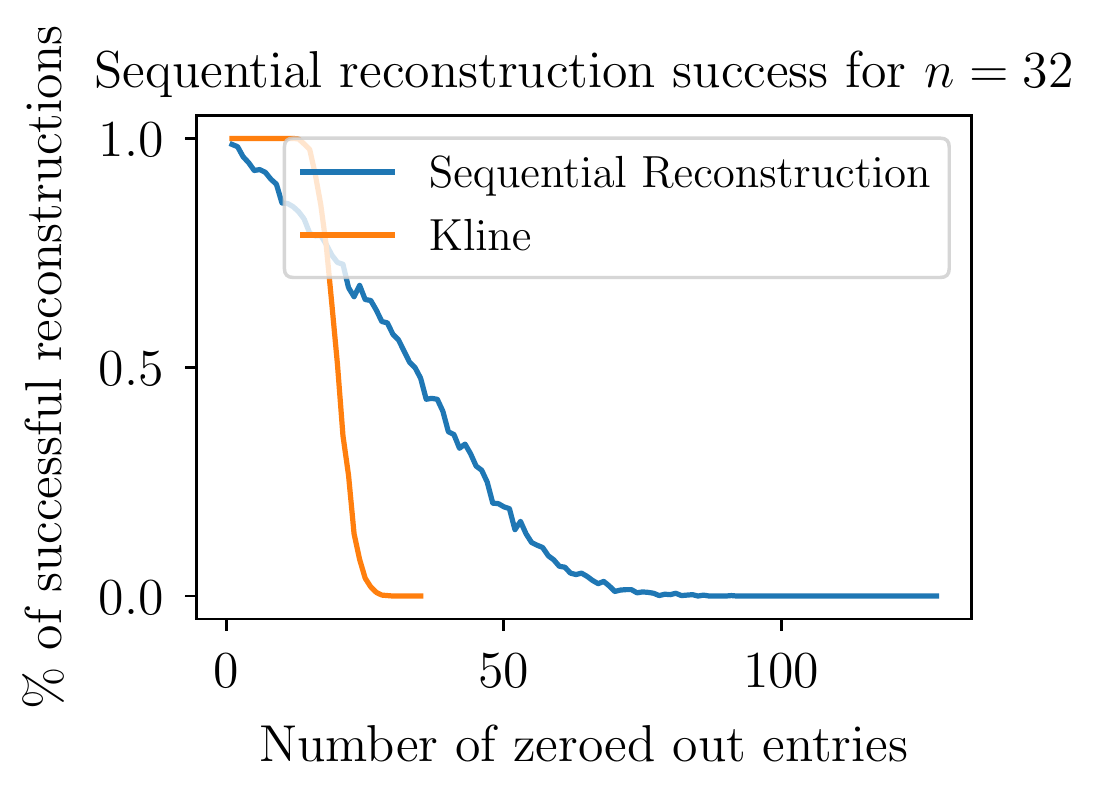}
  \caption{Sequential reconstruction compared to Kline's method for $n=32$.}
  \label{fig:seq-rec-32}
\end{figure}

As we can see the accuracy of HCP remained always above $95\%$. This motivated
the introduction of another reconstruction routine called \textit{sequential
  reconstruction}. Here matrices are reconstructed by successively choosing to
complete the entry corresponding to the HCP. A new prediction is requested from
the model every time an entry is reconstructed. Figure~\ref{fig:seq-rec-32}
shows the performance of this reconstruction for $n=32$.

We can see that the usage of the model in this sequential reconstruction
scenario vastly increases its performance, even surpassing Kline's method.

\section{Conclusions}

In this paper we have shown that DL methods can in fact learn how to reconstruct
Hadamard matrices and that, the success in this task, is heavily dependent on the
usage of GDL techniques. We have shown how to construct an
$S_n^2$-equivariant model and tested it, with great successes, against other
state-of-the-art NNs.

Using the trained models in a one-shot reconstruction scenario yielded, for
$n=24, 28 \textup{ and } 32$, worse results than the current best method in
scientific literature. However, we have shown that the models can be used in
other ways to increase their performance. In our case it was a simple sequential
reconstruction but we expect that the same models could do even better if they
were integrated in a search procedure (\textit{e.g}, beam search).

Lastly, we have seen that our model can effectively generalize to unseen
equivalent classes needing only a few representatives to do so.

\section{Future work}

Recall that, even though we have successfully created an equivariant model to
the actions of $S_n^2$, we did nothing regarding row/column negations. A next
step would be to try and create a model that, in addition, would also be
equivariant to negations thus completely eliminating the need for data
augmentation.

Since our models were successful in the reconstruction task we would like to
pursue more ambitious paths. We could try error correction instead of
reconstruction. Here instead of zeroing out some entries we would flip them and
have the network predict the positions of the errors. We would also like to try
and use said network after some local search method (\textit{e.g}, hill
climbing) to see if the jump from a local optima to a Hadamard matrix could be
made.

Alternatively, we could formulate the search for Hadamard matrices as
reinforcement learning instances and use our networks as the backbone of our
model.

Additionally, we would like to test if, besides generalization to other
equivalent classes, we could achieve generalization across different sizes. Our
hopes of succeding are here build upon the large success that message passing
neural networks have had in this type of generalization. Such is the case
in~\cite{gilmer2017neural} and~\cite{NIPS2017_d9896106} where the authors
observe that their networks generalize well to instances larger than those seen
during training. A positive result here would be of utmost importance since we
could train neural networks in regimes where thousands of matrices are known and
then try to use them for the discovery of unknown classes of larger orders.


\printbibliography

\pagebreak

\appendix

\section{Group theory and equivariance}
\label{sec:grou-appendix}

We cover a few mathematical concepts often appearing in literature related to
GDL.

\begin{definition}
  A \textit{group} is a set $G$ together with an operation $\circ : G \times G
  \mapsto G$ obeying the following conditions:

  \begin{itemize}
  \item If $g, h, f \in G$ then $(g \circ h) \circ f = g \circ (h \circ f)$;
  \item There exists an element $e$, called \textit{identity}, such that $g
    \circ e = e \circ g = g$;
  \item For all elements $g \in G$ there is a unique element $g^{-1} \in G$, often called the
    \textit{inverse}, such that $g \circ g^{-1} = g^{-1} \circ g = e$; 
  \item If $g, h \in G$ then $g \circ h \in G$.
  \end{itemize}

  When no confusion arises we often refer to the group $(G, \circ)$ as just
  group $G$.
\end{definition}

\begin{definition}
  A \textit{group action} of group $G$ on a set $\Omega$ is a mapping $act: G
  \times \Omega \mapsto \Omega$ such that $act(g_1, act(g_2, \omega)) = act(g_1
  \circ g_2, \omega)$ for $g_1, g_2 \in G$ and $\omega \in \Omega$.
\end{definition}

\begin{definition}
  We say that a function $f: \Omega_1 \mapsto \Omega_2$ is \textit{equivariant}
  to the actions of group $G$ if for all $g \in G$ and for all $\omega \in
  \Omega_1$ we have that $f(act_1(g, \omega)) = act_2(g, f(\omega))$. Where
  $act_1$ and $act_2$ are group actions of $G$ on sets $\Omega_1$ and $\Omega_2$
  respectively.
\end{definition}

This definition is perhaps easier to understand by looking at the commutative
diagram below.

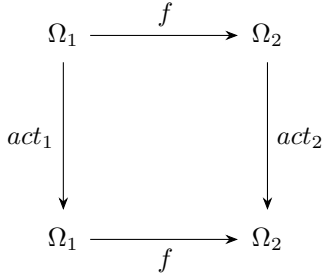
\begin{figure}[h]
  \centering
  \begin{tikzpicture}[
    >=Stealth,
    shorten >=1pt,
    auto,
    node distance=2 cm,
    scale = 1,
    transform shape,
    state/.style={circle,inner sep=2pt}]

    \node[state] (q1) {$\Omega_1$};
    \node[state] (q2) [below=of q1]{$\Omega_1$};
    \node[state] (q3) [right=of q1]{$\Omega_2$};
    \node[state] (q4) [below=of q3]{$\Omega_2$};

    \path[->]
    (q1) edge [left] node {$act_1$} (q2)
    (q1) edge [above] node {$f$} (q3)
    (q2) edge [below] node {$f$} (q4)
    (q3) edge [right] node {$act_2$} (q4);
  \end{tikzpicture}
  \label{fig:equi_diagram}
  \caption{Diagram illustrating equivariance}
\end{figure}

\begin{definition}
  We say that a function $f: \Omega_1 \mapsto \Omega_2$ is \textit{invariant} to
  the actions of group $G$ if for every $g \in G$ and every $\omega \in
  \Omega_1$ we have that $f(act_1(g, \omega)) = f(\omega)$.
\end{definition}

\begin{definition}
  A permutation of size $n$ is a bijective function $\sigma : \{1, \dots, n\}
  \mapsto \{1, \dots, n\}$. The set of all permutations of size $n$ together
  with function composition forms a group to which we will refer to as $S_n$.
  The identity element of $S_n$ is simply the function $e(i) = i \,\forall i \in
  \{1, \dots n\}$. From $S_n$ we can build the group $S_n^2$ whose elements have
  the form $(\sigma_1, \sigma_2)$ with $\sigma_1, \sigma_2 \in S_n$. The binary
  operation of $S_n^2$ is defined as $(\sigma_1, \sigma_2) \circ (\sigma_3,
  \sigma_4) = (\sigma_1 \circ \sigma_3, \sigma_2 \circ \sigma_4)$ and the
  identity element is $(e, e)$.
\end{definition}

Let $M$ be some square matrix of size $n \times n$. We can define the group
actions of $S_n^2$ on $M$ as follows:
\begin{align}
  \label{eq:group_action}
  \left[act((\sigma_1, \sigma_2), M)\right]_{(i, j)} = \left[M\right]_{(\sigma_1^{-1}(i), \sigma_2^{-1}(j))}
\end{align}

Where $[M]_{(i, j)}$ denotes the entry $(i, j)$ of the matrix. Informally this
means that an element $(\sigma_1, \sigma_2)$ of $S_n^{2}$ acts on matrices by
applying $\sigma_1$ to its rows and $\sigma_2$ to its columns, \textit{i.e},
after the group action the $i$-th row/column in the matrix will now have the
values that were present in the $\sigma_{1/2}^{-1}(i)$-th row/column of the
original matrix. Equivalently we could say that the $i$-th row/column in the
matrix will moved into the $\sigma_{1/2}(i)$-th row/column of $act((\sigma_1,
\sigma_2), M)$.


\section{Message passing network training}

Our equivariant models where all trained using four message passing layers
following equation~\ref{eq:layer-formula}. The invariant aggregator chosen was
$\sum$ and we actually used $\psi_{\textup{r}} = \psi_{\textup{c}} := \psi$.
These trainable functions corresponded to simple dense functions, \textit{i.e}:
\begin{align}
  \psi(x) = \rho(Wx)
\end{align}

where $W$ is a matrix with trainable weights and $\rho$ is a non linearity, in
our case $\tanh$. In each layer $i$ its trainable function $\psi$ had output
dimension $2^{i+2}$, \textit{i.e}, $\psi_1: \{-1, 0, 1\}^2 \mapsto \left] -1, 1
\right[^8$, $\psi_2 : \left] -1, 1 \right[^{16} \mapsto \left] -1, 1
\right[^{16}$, $\psi_3 : \left] -1, 1 \right[^{32} \mapsto \left] -1, 1
\right[^{32}$ and $\psi_4 : \left] -1, 1 \right[^{64} \mapsto \left] -1, 1
\right[^{64}$.

As a final classifier we used a multi-layer perceptron with 400, 200, 200 and 1
units in the first, second, third and fourth layers respectively. Once again all
activation function were $\tanh$.

On each epoch the NNs processed 50 batches each with 150 matrices.
We stopped the training process if no improvements where seen after 10 epochs.
Usually the training stopped after about 50 epochs.

We tested two loss functions. $i)$ mean squared error between the network's
output and the target prediction. $ii)$ binary cross-entropy between the neural
networks output and the target predictions. However, because both the predictions
and the targets have values between $-1$ and $1$, we first scaled everything to
the interval $]0, 1[$ using the mapping $x \mapsto (x + 1)/2$. This was
motivated by the fact that matrix completion is in fact a classification
problem.

No significant differences where observed between these two functions. All the
results related to the message passing models presented in this paper are from
networks trained using mean squared error as their loss.

\section{Convolutional neural network training}

For our convolutional neural networks we opted for a structure with 4 layers.
The first three layers each had 32 filters of size 3. The final layer was still
a convolutional layer but with a kernel of size 1. The activation function for
all layers was $\tanh$. They where trained on batches of size 300 and 200
batches in each epoch. We stopped training after 10 epochs without improvement
to the validation loss.

For $n=4$ and $n=8$ this architecture obtained worst results than our method,
but still comparable to Kline's results. However once we reached $n=12$ the
performance fell dramatically. To mitigate this we tried several things:

\begin{itemize}
\item Increase the number of filters in each layer from $32$ to $64$;
\item Increase both the number and size of the filter from $32$ and $3$ to $64$
  and $5$ respectively;
\item Increase both the number and size of the filter from $32$ and $3$ to $128$
  and $5$ respectively;
\item Increase both the number and size of the filter from $32$ and $3$ to $128$
  and $5$ respectively. Added one more layer;
\end{itemize}

None of these changes increased the performance of the NN.

\section{Transformer training}

We also tried an architecture based on just the encoder part of the transformer
as it is presented in~\cite{vaswani2017attention}. We replaced the positional
encoding using $\sin$ and $\cos$ functions for our own that used the exact
position of the element in the matrix. That is, the matrix was flattened into a
sequence of its $\{-1, 1\}$-entries and then each element in the sequence was
augmented with its position in the matrix. For example, the matrix:
\begin{align}
  \begin{bmatrix}
    a & b \\
    c & d
  \end{bmatrix}
\end{align}
after the application of the positional encoder, would be transformed into:
\begin{align}
  \left[ [a, 0, 0], [b, 0, 1], [c, 1, 0], [d, 1, 1] \right]
\end{align}

After the positional encoding we used a trainable embedding and then composed
several encoder structures exactly as defined in~\cite{vaswani2017attention}.
Finally the output of the last encoder was passed to a final classifier.

In each model we can experiment with several different parameters:

\begin{itemize}
\item The embedding, which, in our case, was a simple MLP;
\item The structure of the feed forward network inside each of the encoder
  blocks, again we chose MLPs;
\item The number of attention heads in each multi-head attention component of
  the encoder blocks;
\item The number of encoder block;
\item The structure of the final classifier. Again we opted for an MLP.
\end{itemize}

For the embedding we tried several different MLPs. We started with a simple
dense layer and tested it with dimensions $64$ and $128$. Because this proved
unsuccessful we tried to increase the size of this embedding to an MLP with four
layers of sizes $16$, $32$, $64$, $128$ respectively. In addition we tried to
use both $\tanh$ and $\textup{relu}$ activation functions.

We experimented with a different number of encoder blocks ranging from just one
to four. In each individual  experiment the encoder blocks had the same
structure, \textit{i.e}, they had the same number of attention heads and the
feed forward NNs had the same structure.

For the MLPs in each encoder block we started by using simple dense layers,
first with dimension $64$ and then $128$. We then tried architectures with two
layers, each with dimension $64$, and then each with dimension $128$. Finally we
tested with three layers, first each with dimension $64$, and then $128$. Again
both $\tanh$ and $\textup{rule}$ were used.

The totally of our tests consisted on the combination of the previous parameters
and none had success in the reconstruction task.

\section{Extra visualizations}

In this section we show the EMPM predictions compared to the ground truths of
the data sets.

\begin{figure}[h]
  \centering
\begin{subfigure}{0.5\textwidth}
  \includegraphics[width=\linewidth]{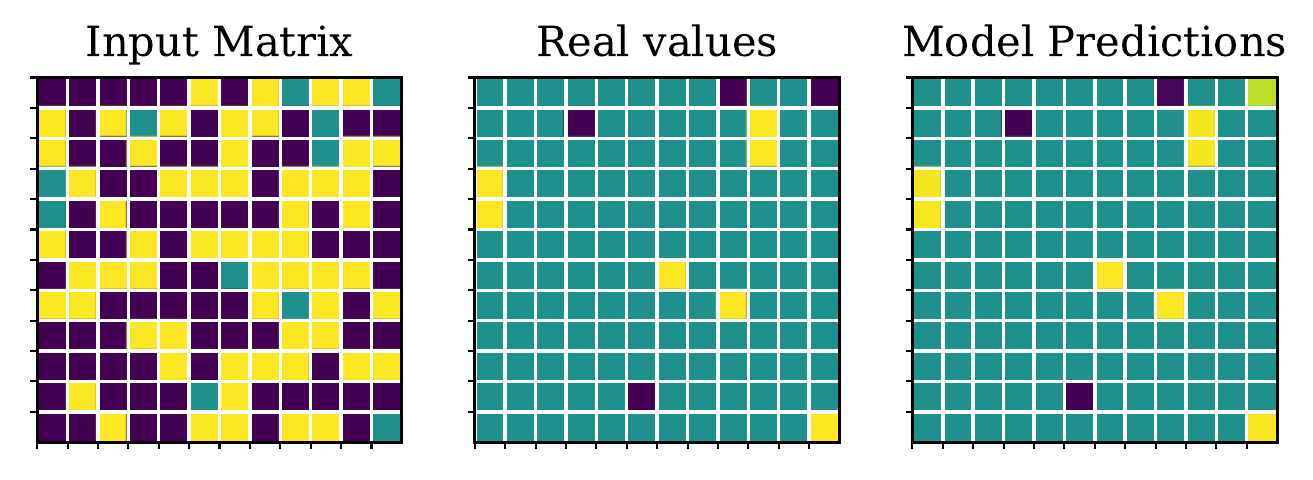}
  \caption{Incorrect decoding}\label{fig:size-12-extra-1}
\end{subfigure}

\medskip
\begin{subfigure}{0.5\textwidth}
  \includegraphics[width=\linewidth]{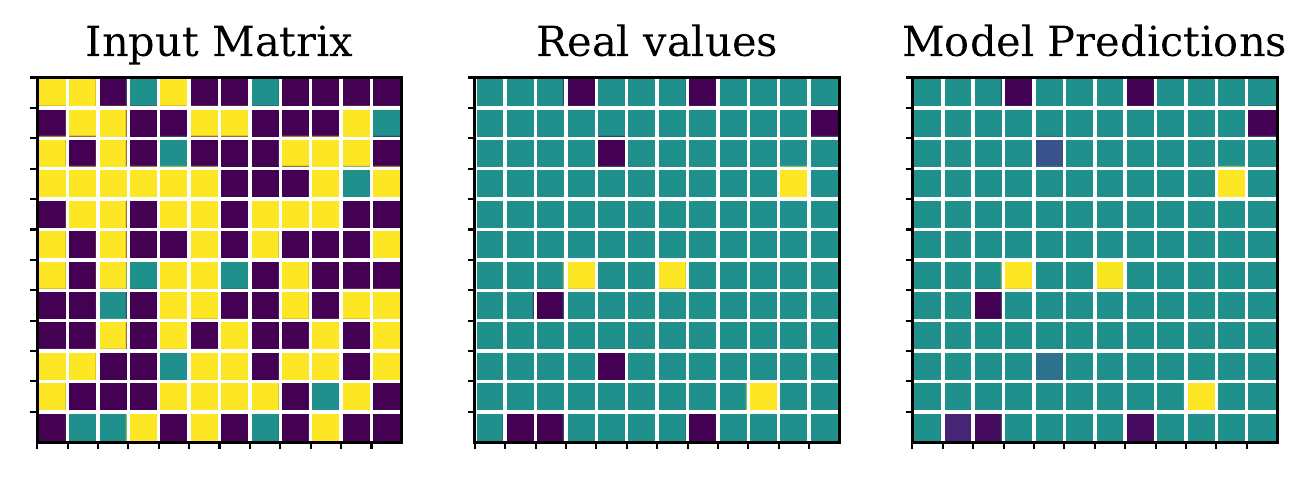}
  \caption{Correct decoding}\label{fig:size-12-extra-2}
\end{subfigure}

\medskip
\begin{subfigure}{0.5\textwidth}
  \includegraphics[width=\linewidth]{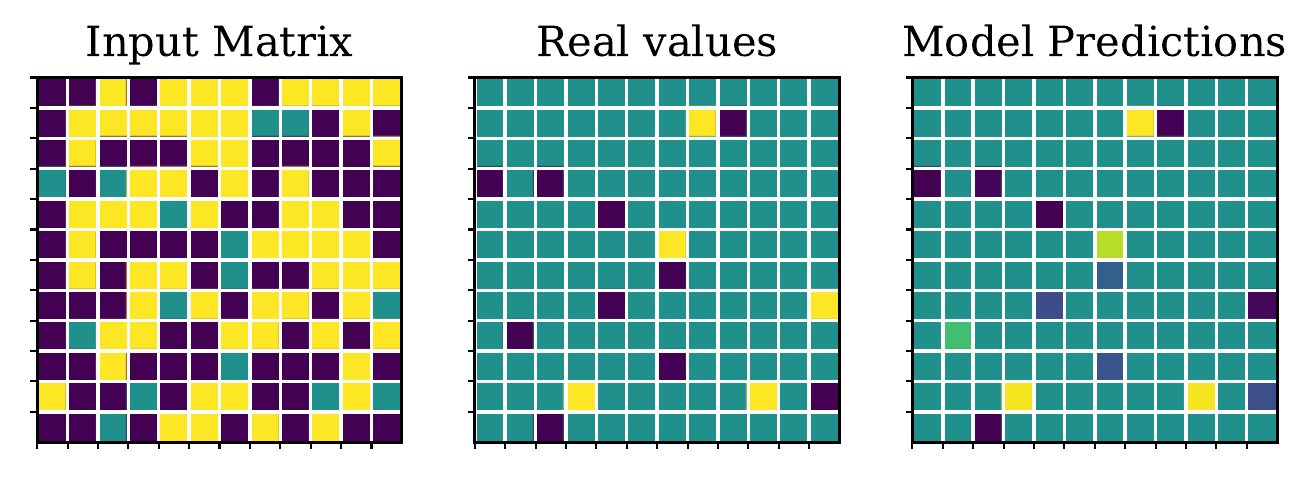}
  \caption{Incorrect decoding}\label{fig:size-12-extra-3}
\end{subfigure}
\caption{Examples of model predictions for $n=12$. The bright yellow predictions
  represent $+1$ values whereas the dark blue colors represent $-1$ values. The
  leftmost matrices represent the matrices with zeroed out entries that need
  completion. The middle matrices represent the target prediction. The rightmost
  matrices represent the EMPM predictions.}\label{fig:size-12-extra}
\end{figure}

Figure~\ref{fig:size-12-extra} shows the model predictions compared to the
ground truths for the model trained for $n=12$. We can see that, using one-shot
reconstruction, in Figure~\ref{fig:size-12-extra-1} the model would make an
incorrect decoding due to the prediction in the top-left corner. Notice also
that that prediction is one of the \textit{weakest} predictions. This is a good
example of the power of using the models for~\textit{sequential reconstruction}.
Probably, if the model was allowed to correct a few of the other entries and
make new predictions after each completion, it would eventually conclude that
the top-left corner should be completed with $-1$.

\end{document}